%% file: main.tex
\documentclass{article}
\usepackage{iclr2027_conference,times}
\iclrfinalcopy
\input{math_commands.tex}

\usepackage[T1]{fontenc}
\usepackage{algorithm}
\usepackage{algorithmic}
\usepackage{hyperref}
\hypersetup{hidelinks}
\usepackage{url}
\usepackage{booktabs}
\usepackage{graphicx}
\usepackage{colortbl}
\usepackage{array}
\usepackage{arydshln}
\usepackage{multirow}
\usepackage{titletoc}
\newcommand{\method}{\textsc{FDC}}

\newcommand{\normfeat}{\operatorname{norm}}
\newcommand{\Std}{\operatorname{Std}}
\newcommand{\pct}[2]{\ensuremath{#1\!\pm\!#2}}
\newsavebox{\fdctablebox}
\usepackage{acro}
\DeclareAcronym{Ours}{
  short = \texttt{FDC},
  long  = \textbf{F}eature \textbf{D}istribution \textbf{C}alibration}
  
\usepackage{xspace}
\renewcommand{\eqref}[1]{(\ref{#1})}
\title{Distribution-Conditioned Task Routing \\
for Class-Incremental Learning}
\author{\textbf{Longhuan Xu$^{1,*}$, Zhipeng Zhou$^{2,*}$, Wei Ji$^{1}$,}\\
\textbf{Chunyan Miao$^{2}$, Peilin Zhao$^{3,\dagger}$, Lijun Zhang$^{1,\dagger}$}\\[0.5em]
\normalfont $^{1}$Nanjing University\\
\normalfont $^{2}$Nanyang Technological University\\
\normalfont $^{3}$Shanghai Jiao Tong University\\[0.4em]
\normalfont\small $^{*}$Joint first authors.\quad $^{\dagger}$Corresponding authors.}
\date{}
\begin{document}
\maketitle
% Clear the conference publication header for this preprint.
\lhead{}
\input{sections/abstract}
\input{sections/introduction}
\input{sections/related}
\input{sections/motivation}
\input{sections/design}
\input{sections/experiments}
\input{sections/limitations}
\input{sections/conclusion}
\input{sections/ai}
\begingroup
\raggedright
\bibliographystyle{iclr2027_conference}
\bibliography{references}
\endgroup

\clearpage
\appendix
\raggedbottom

\setcounter{tocdepth}{2}
\startcontents[appendix]
\begin{center}
    {\Large\bfseries Appendix}
\end{center}
\vspace{0.5em}
\printcontents[appendix]{5}{1}{}

\newpage
\input{sections/appendix}

\end{document}

%% file: math_commands.tex
\usepackage{amsmath,amsfonts,bm}

\def\eqref#1{equation~\ref{#1}}
\def\1{\bm{1}}

\DeclareMathAlphabet{\mathsfit}{\encodingdefault}{\sfdefault}{m}{sl}
\SetMathAlphabet{\mathsfit}{bold}{\encodingdefault}{\sfdefault}{bx}{n}

%% file: sections/abstract.tex
% !TEX root = ../main.tex
\begin{abstract}
Parameter-efficient adaptation enables continual learners to acquire task-specific knowledge through compact model updates while maintaining strong within-task performance. However, class-incremental inference requires each input to be classified among all classes seen so far without access to its task identity. For learners equipped with task-specific parameter-efficient modules, this introduces a critical task-routing challenge beyond catastrophic forgetting. We study post-hoc task routing without retraining the learner or introducing a separately trained router. Such training-free inference-time calibration remains comparatively underexplored in parameter-efficient class-incremental learning. We identify three sources of routing error---feature-level, task-level, and class-level misalignment---and propose Feature Distribution Calibration (\method{}). Its three components address these misalignments: Task Subspace Filtering (TSF) suppresses feature components outside each task's principal subspace, Residual Likelihood Calibration (RLC) evaluates the typicality of its subspace residual, and Prototype Affinity Calibration (PAC) measures compatibility with the task's class prototypes. Experiments demonstrate plug-and-play applicability to eight parameter-efficient class-incremental methods using a shared encoder. With one component configuration selected per method across all five benchmarks, \method{} improves final accuracy in all 40 method--dataset pairs by 4.39 percentage points on average. Enabling all components improves 35 of the 40 pairs, with an average gain of 4.45 points. When applied to a simple baseline, \method{} achieves strong overall performance.
\end{abstract}

%% file: sections/introduction.tex
% !TEX root = ../main.tex
\section{Introduction}
\label{sec:intro}
Pre-trained models (PTMs)~\citep{zhou2024ptmsurvey} have achieved remarkable success and substantially reshaped the paradigm of continual learning (CL). Traditional CL methods~\citep{khademinori2025ahr,wang2026pronc} often adapt a large portion of model parameters while explicitly addressing catastrophic forgetting, whereas PTMs enable more parameter-efficient alternatives, such as prompt tuning~\citep{wang2022l2p,hong2025rainbowprompt} and Low-Rank Adaptation (LoRA)~\citep{hu2022lora}, to preserve previously acquired knowledge while maintaining strong task-specific adaptation. Consequently, LoRA-based class-incremental learning (CIL) has attracted increasing attention in recent years.  

\begin{figure}[!t]
\centering
\includegraphics[width=\linewidth]{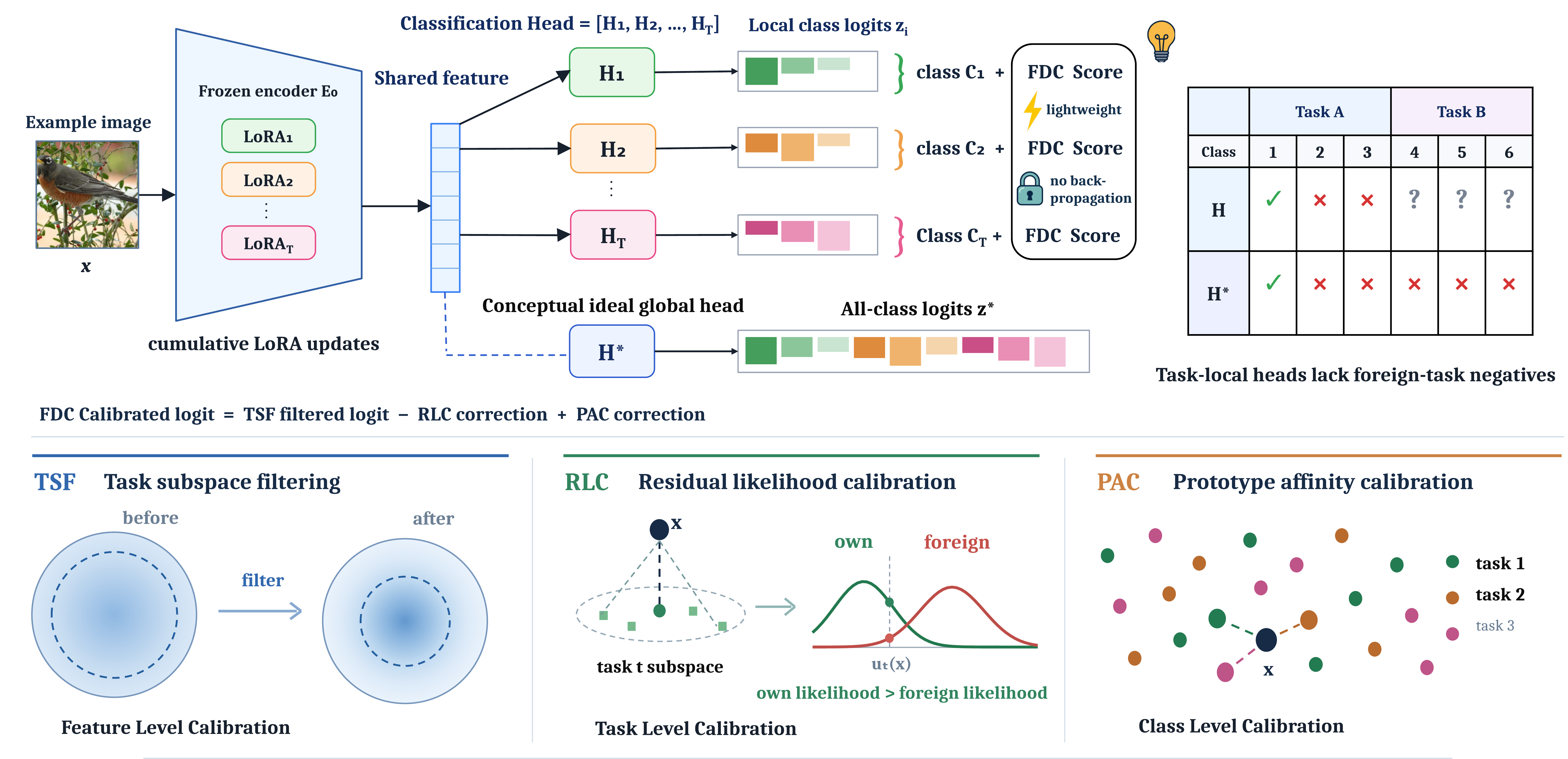}
\caption{Post-hoc task routing with \method{}. Top: task-local heads lack foreign-task negatives; $H^*$ illustrates an ideal global classifier (Section~\ref{sec:motivation}). Bottom: TSF, RLC, and PAC calibrate routing through feature support, task membership, and class affinity, respectively, without additional training.}
\label{fig:overview}
\end{figure}

Existing LoRA-based CIL methods~\citep{wu2025sdlora,liu2025sub,qiu2026splitlora} have devoted substantial effort to mitigating catastrophic forgetting as the label space continually expands, achieving competitive performance across a variety of benchmarks. However, we highlight another complementary yet critical challenge: \emph{task routing}, i.e., identifying the appropriate task-specific knowledge for an input when task identity is unavailable at inference time. This challenge becomes particularly important because, as empirically demonstrated in Figure~\ref{fig:motivation-gap}, within-task discrimination is largely preserved, whereas discrimination across tasks remains substantially impaired. This suggests that routing, rather than within-task recognition itself, can become a key bottleneck in class-incremental inference. While task routing itself is not new~\citep{kim2022theory,wang2023hide}, existing solutions typically rely on learned routers~\citep{wang2025sema}, out-of-distribution (OOD)-aware continual training~\citep{jia2026god}, or prototype- and distribution-based prediction mechanisms~\citep{lin2024tpl}, all of which introduce additional training requirements or modify the original learning procedure. In contrast, whether task routing can be effectively improved \emph{after} a LoRA-based CIL model has already been trained remains largely unexplored.

In this paper, we investigate this practical yet challenging setting, termed \emph{post-hoc task routing} for LoRA-based CIL. Given an already trained CIL learner, our goal is to improve task routing without updating the encoder or classifier parameters and without training an additional router. Meanwhile, the additional storage and computational overhead should remain modest to preserve the efficiency benefits of parameter-efficient adaptation. These constraints require us to extract discriminative routing evidence directly from the established feature space. To this end, we analyze routing errors from a coarse-to-fine perspective and identify three distinct forms of misalignment: \emph{feature-level}, \emph{task-level}, and \emph{class-level} misalignment. We empirically verify their prevalence across existing benchmarks in Sec.~\ref{sec:motivation}. At the feature level, residual components outside a task's principal support tend to make positive logit contributions for foreign-task inputs, while contributing little or even negatively to own-task inputs. At the task level, own-task and foreign-task inputs exhibit distinct residual-energy distributions, providing useful evidence for task membership. At the class level, individual class prototypes preserve fine-grained structure that can be obscured by task-level means.

Building on these observations, we propose a unified post-hoc calibration framework, \method{} (Figure~\ref{fig:overview}), to systematically address the three levels of routing misalignment. Specifically, \emph{Task Subspace Filtering} (TSF) suppresses feature components outside each task's principal subspace before applying its classifier, thereby mitigating feature-level misalignment. \emph{Residual Likelihood Calibration} (RLC) evaluates whether the residual geometry of an input is more typical of the candidate task or of foreign tasks, providing task-level evidence. Finally, \emph{Prototype Affinity Calibration} (PAC) measures the input's affinity to each task's class prototypes relative to its training distribution, complementing routing decisions with class-level evidence. Together, these components progressively calibrate routing from feature support to task membership and class affinity, while contributing to the scores of all task classification heads without additional gradient-based training.

Our contributions are threefold:
\begin{itemize}
\item We investigate a practical yet underexplored setting, \emph{post-hoc task routing} for LoRA-based CIL, where routing is improved after training without modifying the learned encoder or classifier. We further identify and empirically characterize three complementary sources of routing error: feature-, task-, and class-level misalignment.
\item We propose \method{}, a lightweight post-hoc calibration framework that jointly addresses these misalignments through feature support projection, task-level residual evidence, and class-prototype affinity. \method{} requires neither additional gradient-based optimization nor a replay buffer, while keeping the learned encoder and classifier parameters fixed.
\item Across five benchmarks and three training seeds, \method{} improves Cumulative-LoRA by 3.58--10.88 percentage points. Moreover, plug-and-play evaluation on eight additional CIL learners yields positive average gains across all 40 method--dataset combinations, with an overall improvement of 4.39 percentage points using a single component configuration per learner across the five benchmarks.
\end{itemize}

%% file: sections/related.tex
% !TEX root = ../main.tex

\section{Related Work}
\label{sec:related}

\subsection{Parameter-Efficient Class-Incremental Learning}

CIL requires recognizing all previously encountered classes without task
labels at inference. Pretrained vision transformers provide transferable representations
\citep{dosovitskiy2021vit}, which continual learners adapt through compact prompts, adapters, or
low-rank updates. RainbowPrompt aggregates task-specific prompts through learned
transformations and alignment \citep{hong2025rainbowprompt}, while adapter-based
approaches explore task-specific and shared adaptation \citep{wang2025tuna,zhao2026rsiat}.
Among these approaches, LoRA parameterizes weight updates with low-rank factors \citep{hu2022lora}.
InfLoRA and SplitLoRA constrain update spaces to manage interference
\citep{liang2024inflora,qiu2026splitlora}; SD-LoRA decouples update magnitude and direction
\citep{wu2025sdlora}; and LoDA separates knowledge-sharing and task-specific subspaces
\citep{he2026loda}. Complementary strategies regularize low-rank updates
\citep{zheng2026ewclora}, allocate capacity proactively \citep{wang2025plan}, or construct
drift-resistant representations \citep{liu2025sub}. \method{} complements these training strategies by using feature statistics to calibrate competition among task heads, while keeping the learned encoder and classifier parameters fixed.

\subsection{Task Routing and Classifier Calibration}

\citet{kim2022theory} connect within-task recognition and task inference to OOD detection;
HiDe-Prompt further separates within-task, task-identity, and task-adaptive prediction
\citep{wang2023hide}. Modular learners address task inference through learned adapter weights
(SEMA)~\citep{wang2025sema}, training-free self-refined retrieval (MOS)~\citep{sun2025mos}, or
entropy-based selection over candidate-specific predictions (GR-LoRA)~\citep{lin2026grlora}.
\method{} instead calibrates competition among heads sharing one adapted representation.

Weight Aligning corrects classifier bias by rescaling weights without further optimization
\citep{zhao2020wa}; SLCA optimizes the classifier using sampled class-wise features
\citep{zhang2023slca}. T-CIL targets confidence calibration through temperature optimization
with perturbed memory exemplars and new-task validation data~\citep{hwang2025tcil}.
\method{} targets cross-task classification errors; Section~\ref{sec:plugin} tests its
compatibility with classifier alignment through InfLoRA+CA.

Feature statistics also support inference: FeCAM uses class means and covariances for
Mahalanobis classification~\citep{goswami2023fecam}, and AnaCP analytically learns a contrastive
projection and classifier~\citep{momeni2025anacp}. TPL predicts tasks using likelihood ratios
with task-specific models and replay-based estimates~\citep{lin2024tpl}; ViM combines subspace
residuals with logits for OOD detection~\citep{wang2022vim}. GOD instead shapes feature geometry
during training through ETF and ArcFace losses~\citep{jia2026god}. \method{} combines subspace support, residual typicality, and prototype affinity to calibrate
existing heads, using compact statistics without replay or additional gradient-based training.

%% file: sections/motivation.tex
% !TEX root = ../main.tex
\section{Preliminary and Motivation}
\label{sec:motivation}
\subsection{Preliminary}
Consider $T$ tasks with training sets $\mathcal D_1,\ldots,\mathcal D_T$ and mutually disjoint class sets $\mathcal C_1,\ldots,\mathcal C_T$. Let $E_0$ be the frozen pretrained encoder and $E_t$ the encoder after learning task $t$. For each adapted matrix, low-rank updates accumulate as $W_{\rm enc}^{(t)}=W_{\rm enc}^{(0)}+\sum_{k=1}^t B_kA_k$. Task $t$ learns a head $g_t(h)=W_th+b_t$ for its classes. Its current-task training minimizes empirical classification loss:
\begin{equation}
 \hat\theta_t\in\arg\min_{\theta_t}
 \frac{1}{|\mathcal D_t|}\sum_{(x,y)\in\mathcal D_t}
       \ell(g_t(E_t(x)),y),
 \label{eq:intro-erm}
\end{equation}
where $\theta_t$ comprises the new LoRA increment and current head. Historical classifier blocks remain fixed. At final inference, all heads share $E_T$, and prediction spans all seen classes:
\begin{equation}
 z_t(x)=W_tE_T(x)+b_t,\qquad
 (\hat t,\hat c)=\arg\max_{t,\;c\in\mathcal C_t}z_{t,c}(x).
 \label{eq:raw}
\end{equation}

The local argmax in the selected head yields the global class prediction; the maximum logit indicates the head's preferred class, but training on $\mathcal D_t$ does not directly constrain its response to another task.

\subsection{Motivation}
\paragraph{Task routing is a key challenge.}
\begin{figure}[!t]
\centering
\newsavebox{\motivationrightbox}
\setlength{\abovecaptionskip}{3pt}
\setlength{\belowcaptionskip}{0pt}
\def\motivationtablerows{\toprule Dataset & Accuracy change (pp) \\
\midrule
CIFAR-100 & $+0.35\pm0.85$ \\
ImageNet-A & $-0.02\pm0.34$ \\
ImageNet-R & $-0.79\pm0.34$ \\
CUB-200 & $-0.54\pm1.23$ \\
OmniBenchmark & $+0.11\pm0.23$ \\
\bottomrule}
\sbox{\motivationrightbox}{%
\begin{minipage}[t]{0.45\linewidth}
\vspace{0pt}\centering\small
\setlength{\tabcolsep}{3pt}
\renewcommand{\arraystretch}{0.95}
\begin{tabular}{@{}lc@{}}\motivationtablerows\end{tabular}
\begingroup
\expandafter\def\csname @captype\endcsname{table}
\caption{Accuracy change after LS.}
\label{tab:logit_standardization}
\endgroup

\vspace{3pt}
\begin{tabular}{@{}lcc@{}}
\toprule
TSF & RLC & PAC \\
\midrule
$\mathcal O(Td^2)$ & $\mathcal O(Td^2)$ & $\mathcal O(Cd)$ \\
\bottomrule
\end{tabular}
\begingroup
\expandafter\def\csname @captype\endcsname{table}
\caption{Extra storage for FDC statistics. $T$: number of tasks; $C$: total number of classes; $d$: classification head input dimension.}
\label{tab:fdc-storage}
\endgroup
\end{minipage}%
}
\begin{minipage}[t]{0.52\linewidth}
\vspace{0pt}\centering
\includegraphics[width=\linewidth]{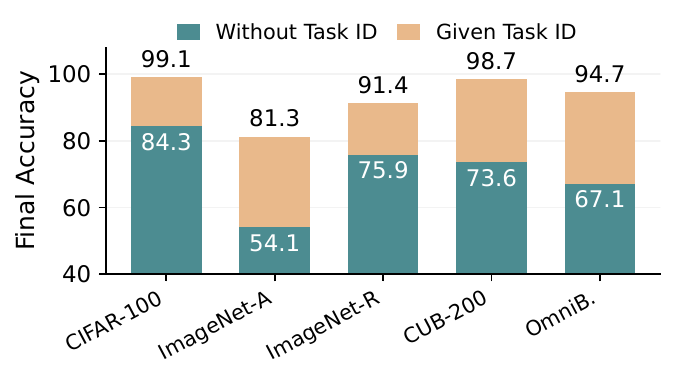}
\caption{Final accuracy comparison between w/ and w/o task identity.}
\label{fig:motivation-gap}
\end{minipage}\hfill
\begin{minipage}[t]{0.45\linewidth}
\vspace{0pt}\usebox{\motivationrightbox}
\end{minipage}
\end{figure}
We compare two prediction settings for the same final model and test inputs. Without Task ID,
the model selects the class with the highest logit across all seen classes; Given Task ID,
it selects the class with the highest logit only within the true task's classes. Both settings
use the same final shared encoder and concatenated classifier heads, with all accumulated LoRA updates active. Providing task identity therefore does not undo any forgetting caused by subsequent
updates, which can still affect recognition within the true task. Nevertheless,
Figure~\ref{fig:motivation-gap} shows a substantial accuracy gap across all five datasets, with
task identity alone correcting around 80\% of errors on average. Thus, much of the required recognition capability remains available in the shared representation but is obscured by cross-head
competition. This makes routing calibration a central priority alongside mitigating catastrophic forgetting during training. We want to clarify that we are re-identifying the task routing misalignment during the post-hoc stage in CIL, rather than presenting this observation as our original contribution.

\paragraph{Limited gains from logit standardization (LS).}
A natural first hypothesis is that the separately trained heads produce logits with different
offsets and scales, potentially biasing their competition. Although the encoder is shared through cumulative LoRA updates, the heads are simply concatenated at inference:
\begin{equation}
 z(x)=\begin{bmatrix}z_1(x)\\ \vdots\\ z_T(x)\end{bmatrix}
 =\begin{bmatrix}W_1\\ \vdots\\ W_T\end{bmatrix}E_T(x)
 +\begin{bmatrix}b_1\\ \vdots\\ b_T\end{bmatrix}.
 \label{eq:concatenated-heads}
\end{equation}
To test this hypothesis, each head's maximum logit is standardized as:
\begin{equation}
 s_t^{\mathrm{LS}}(x)=\frac{s_t^{\rm raw}(x)-\mu_t^s}{\max(\sigma_t,\epsilon)},
 \qquad \hat t_{\rm std}(x)=\arg\max_t s_t^{\mathrm{LS}}(x),
 \label{eq:logit-standardization}
\end{equation}
where $\mu_t^s$ and $\sigma_t$ are the mean and standard deviation of head $t$'s maximum
logit on its own training data, and $\epsilon=10^{-12}$ ensures numerical stability.
This transformation changes head selection while preserving each head's local class prediction.
However, joint centering and scaling changes CIL accuracy by only $+0.35$, $-0.02$, $-0.79$,
$-0.54$, and $+0.11$ percentage points on the five datasets, respectively
(see Table~\ref{tab:logit_standardization}), providing no consistent improvements. Correcting these
score statistics alone therefore appears insufficient for reliable routing. These results
motivate keeping learned logits while supplementing them with evidence about the current
input's compatibility with each task's training distribution.

\begin{figure}[t]
\centering
\begin{minipage}[t]{0.65\linewidth}
\vspace{0pt}
\centering
\includegraphics[width=\linewidth]{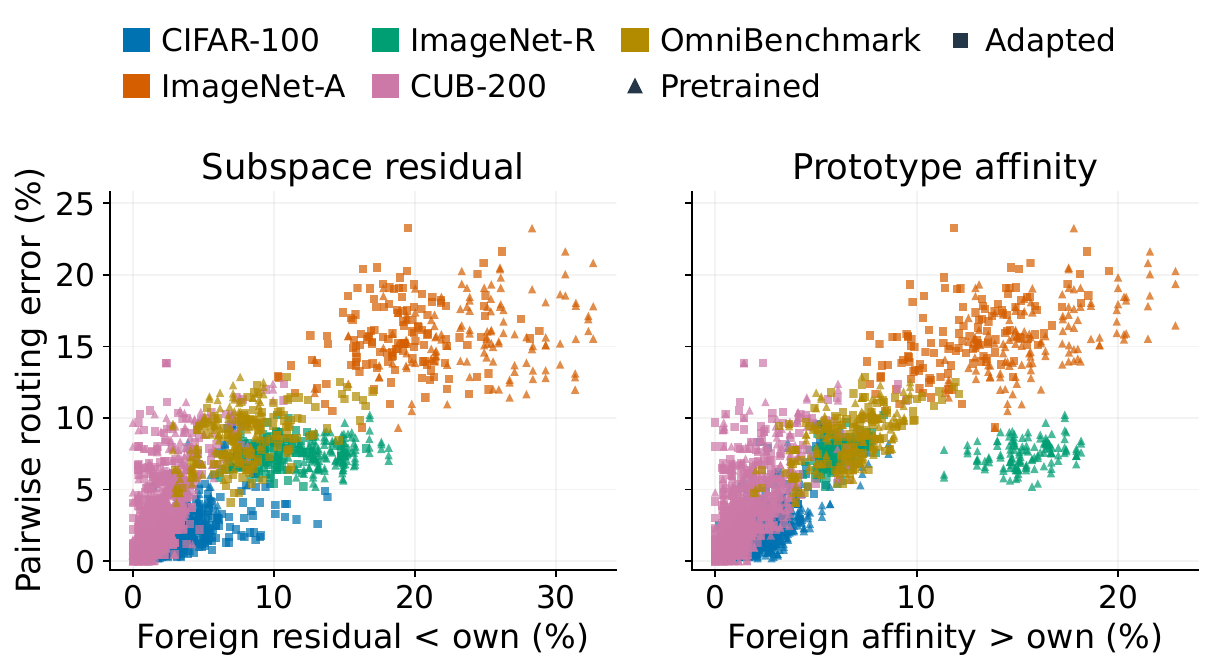}
\caption{Feature similarity versus routing error. Pretrained/adapted use the frozen encoder/encoder after each task. Similarity uses subspace residuals (left) and prototype affinities (right).}
\label{fig:motivation-confusability}
\end{minipage}\hfill
\begin{minipage}[t]{0.33\linewidth}
\vspace{0pt}
\centering
\includegraphics[width=\linewidth]{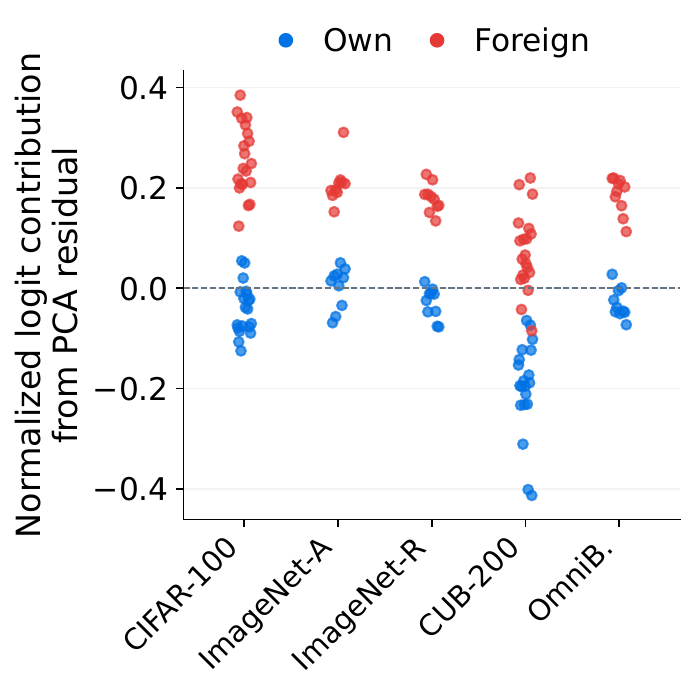}
\caption{PCA-residual logit contributions on own and foreign inputs. Each point denotes one head averaged over its own or foreign test samples.}
\label{fig:routing-osp}
\end{minipage}
\end{figure}

\paragraph{Head input similarity and routing confusion.}
We further empirically identify that head-input similarity is positively correlated with routing confusion. As shown in Figure~\ref{fig:motivation-confusability}, task pairs that are more confusable in terms of subspace residuals (Eq.~\eqref{eq:residual}) or affinities to the class prototypes (Eq.~\eqref{eq:par}) tend to exhibit more routing errors. Here, head-input similarity is assessed by comparing how well an input fits its own task's feature distribution with how well it fits another task's. Subspace residuals measure its fit to each task's principal subspace, while prototype affinities measure its resemblance to each task's class mean prototypes.

\paragraph{Routing errors at three levels.}
To analyze the sources of the error gap in Figure~\ref{fig:motivation-gap}, we compare
the task-locally trained classifier $H$ with a hypothetical jointly supervised
reference $H^*$ serving as the calibration target. Since both settings
use the same model, we assume that $H$ and $H^*$ receive identical inputs.
$H$ trains each head using only within-task class labels. $H^*$ additionally uses
cross-task labels under an all-class classification loss, providing negative
supervision for foreign inputs.

For a foreign class $c\in\mathcal C_{t'}$, $t'\ne t$, write $\bar t$ for
``not task $t$.'' The probability chain rule gives a three-level factorization,
exposing the cross-task supervision available to $H^*$ but absent from $H$:
\begin{equation}
 P(c\mid x)=
 \underbrace{P(\bar t\mid x)}_{\text{not task }t}\;
 \underbrace{P(t'\mid\bar t,x)}_{\text{which task}}\;
 \underbrace{P(c\mid t',x)}_{\text{which class}}.
 \label{eq:cross-task-chain}
\end{equation}
To translate these evidence levels into score corrections, let
$r_t^{\mathrm{feat}},r_t^{\mathrm{task}},r_t^{\mathrm{class}}>0$ denote their
multiplicative calibration factors. Let $s_t,s_t^*$ be the maximum class logits
of $H,H^*$ and $\sigma_t>0$ the head's own-training maximum-logit standard deviation.
We model calibration on the scale $s_t/\sigma_t$ as
\begin{equation}
 \begin{aligned}
 \frac{\exp(s_t(x)/\sigma_t)}{\exp(s_t^*(x)/\sigma_t)}
 &=r_t^{\mathrm{feat}}(x)\,r_t^{\mathrm{task}}(x)\,r_t^{\mathrm{class}}(x),\\
 \overset{\log}{\Longrightarrow}\quad
 \frac{s_t(x)-s_t^*(x)}{\sigma_t}
 &=\log r_t^{\mathrm{feat}}(x)+\log r_t^{\mathrm{task}}(x)+\log r_t^{\mathrm{class}}(x).
 \end{aligned}
 \label{eq:multiplicative-calibration}
\end{equation}
Writing $\Delta_t^\ell(x)=\log r_t^\ell(x)$ for
$\ell\in\{\mathrm{feat},\mathrm{task},\mathrm{class}\}$ and multiplying by $\sigma_t$ gives
\begin{equation}
 s_t^*(x)=s_t(x)-\sigma_t\bigl[
 \Delta_t^{\mathrm{feat}}(x)+\Delta_t^{\mathrm{task}}(x)+\Delta_t^{\mathrm{class}}(x)\bigr].
 \label{eq:three-level-correction}
\end{equation}

We examine these three levels through the following diagnostics.
At the \emph{feature level}, a head's high response to foreign inputs may arise
from directions rarely used by its own-task inputs.
We measure the normalized logit contribution from features outside the task's
principal component analysis (PCA)~\citep{pearson1901pca} subspace
(Appendix~\ref{app:feature-diagnostic}).
Figure~\ref{fig:routing-osp} shows generally positive mean residual contributions on foreign
inputs, versus small or negative contributions on own-task inputs.
This observation motivates TSF (Section~\ref{sec:osp}).

\begin{figure}[!t]
\centering
\includegraphics[width=\linewidth]{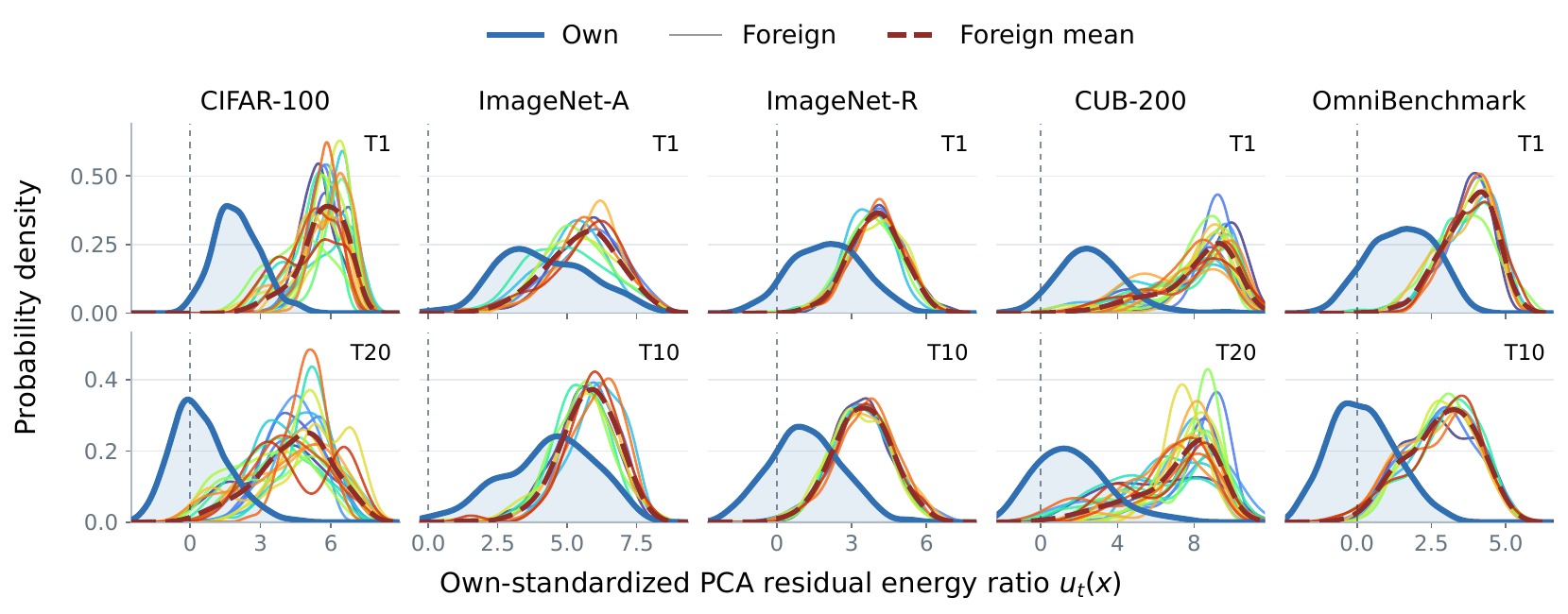}
\caption{Own-standardized PCA residual distributions for the first and last task heads. Dashed curves show the equal-task mean of foreign densities.}
\label{fig:routing-srlr}
\end{figure}

At the \emph{task level}, PCA also provides a measure of how well an input fits a task's
feature distribution. Fitted to samples across the task's classes, its principal subspace
summarizes their dominant feature variation. The fraction of centered feature energy outside
this subspace then measures the input's deviation from that task.
Figure~\ref{fig:routing-srlr} shows that foreign inputs generally have higher residual energy
ratios than own-task inputs, with each task's training statistics used for standardization.
This difference motivates the residual-based task calibration in RLC
(Section~\ref{sec:srlr}).

\begin{figure}[!t]
\centering
\includegraphics[width=\linewidth]{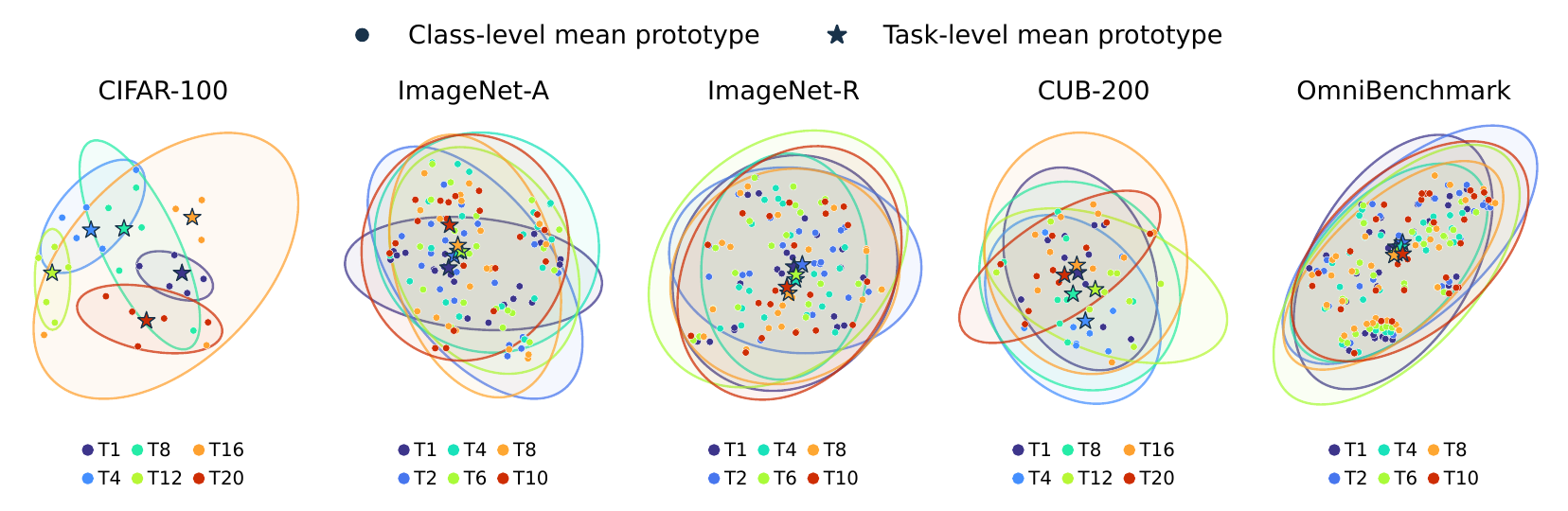}
\caption{Mean prototype geometry across five datasets. Dots are class mean prototypes and stars their task-level mean prototype, jointly embedded with t-SNE.}
\label{fig:routing-par}
\end{figure}

At the \emph{class level}, an input's resemblance to a particular class in another task may help explain why that task's classification head competes with the true task's head. Figure~\ref{fig:routing-par} illustrates how task mean prototypes can cluster closely together while class mean prototypes retain more distinct locations. Averaging classes into a single task prototype can obscure this finer structure, making class-level affinity useful for assessing which task contains a class compatible
with the input.
PAC (Section~\ref{sec:par}) uses this class-level evidence for calibration.

%% file: sections/design.tex
% !TEX root = ../main.tex
\section{Principal Design}
\label{sec:method}
\subsection{Overview}
To address the three levels of routing misalignment in Section~\ref{sec:motivation},
\method{} uses three calibration modules: Task Subspace
Filtering (TSF) for feature-level correction, Residual Likelihood Calibration (RLC)
for task-level correction, and Prototype Affinity Calibration (PAC) for class-level
correction. It calibrates task classification heads without back-propagation or any separately trained routing mechanism by observing the $d$-dimensional input features of task classification heads across
$T$ tasks and $C=\sum_{t=1}^T|\mathcal C_t|$ classes.
Figure~\ref{fig:overview} summarizes the routing problem and the three calibration modules.

For classification head input features $g(x)$, task-level variation is captured by PCA:
\begin{equation}
 U_t=\operatorname{PCA}_{\eta}\bigl(\{g(x):x\in\mathcal D_t\}\bigr),
 \qquad \Pi_t=U_tU_t^\top.
 \label{eq:task-pca}
\end{equation}
PCA centers the features by subtracting their task mean and retains the fewest leading
components explaining at least a fraction $\eta$ of the total variance (Appendix~\ref{app:tsf-pca-sensitivity}).
Class-level structure is captured by averaging the unit-normalized features
$\widehat g(x)=g(x)/\|g(x)\|_2$ within each class:
\begin{equation}
 p_{t,c}=\normfeat\!\left(\frac{1}{|\mathcal D_{t,c}|}
       \sum_{x\in\mathcal D_{t,c}}\widehat g(x)\right),\qquad
 \normfeat(u)=\frac{u}{\|u\|_2}.
 \label{eq:proto}
\end{equation}
Here $\mathcal D_{t,c}$ contains task $t$'s training samples of class $c$. Statistics are collected as tasks are learned.
\subsection{Task Subspace Filtering (TSF)}
\label{sec:osp}
As observed in Figure~\ref{fig:routing-osp}, feature components outside a task's principal
subspace can inflate its logits on foreign inputs. To attenuate these contributions,
TSF uses the adapted-view projection $\Pi_t^t$ from Eq.~\eqref{eq:task-pca}
to transform the task's classifier weights:
\begin{equation}
 \widetilde W_t=W_t-\gamma W_t(I-\Pi_t^t),
 \label{eq:osp}
\end{equation}
where $\gamma\in[0,1]$ controls the attenuation strength (Appendix~\ref{app:tsf-pca-sensitivity}).
This transformation scales the contribution of features orthogonal to the task subspace
by $1-\gamma$, while preserving the contribution within the subspace.
With the learned head bias $b_t$ unchanged, the filtered logits are
$\widetilde z_t(x)=\widetilde W_t E_T(x)+b_t$, giving the task logit (the maximum class logit within head $t$)
$\widetilde s_t(x)=\max_{c\in\mathcal C_t}\widetilde z_{t,c}(x)$.

\subsection{Residual Likelihood Calibration (RLC)}
\label{sec:srlr}
RLC compares an input's subspace residual under the task's own distribution and a
foreign-task reference. For a feature $g(x)$ from either encoder, we use the task
projection $\Pi_t$ from Eq.~\eqref{eq:task-pca} to compute
\begin{equation}
 \rho_t(x)=\frac{\|(I-\Pi_t)(g(x)-m_t)\|_2^2}{\|g(x)-m_t\|_2^2},
 \qquad u_t(x)=\frac{\rho_t(x)-\mu_t^\rho}{\delta_t},
 \label{eq:residual}
\end{equation}
where $m_t$ is the mean training feature, and $\mu_t^\rho$ and $\delta_t$ are the mean and
standard deviation of training residuals, computed separately for each encoder.

We approximate the standardized own-task residuals by $p_O=\mathcal N(0,1)$.
The foreign reference $p_F=\mathcal N(\mu_F,V_F)$ is fitted by evaluating task $t$'s
features $g(x)$ against the stored statistics of earlier tasks.
The resulting likelihood correction is
\begin{equation}
 S_t^{\mathrm{RLC}}(x)=\sigma_t\tanh\!\left(\frac{1}{\tau}
           \log\frac{p_F(u_t(x))}{p_O(u_t(x))}\right),
 \label{eq:llr}
\end{equation}
where $\tau$ is the median absolute log-likelihood ratio over the foreign samples.
RLC subtracts $S_t^{\mathrm{RLC}}(x)$ from the task logit.
A higher foreign likelihood decreases the score, whereas a higher own-task likelihood
increases it.

\subsection{Prototype Affinity Calibration (PAC)}
\label{sec:par}
PAC assesses task compatibility through the closest class prototype, retaining the
class-level structure that a single task prototype can obscure.
For each candidate task $t$, PAC performs a nearest-neighbor search ($K=1$) over
the class prototypes $p_{t,c}$ from Eq.~\eqref{eq:proto}, using cosine similarity.
The similarity to the retrieved prototype
serves as the task affinity, which we standardize using the task's training statistics:
\begin{equation}
 A_t(x)=\max_{c\in\mathcal C_t}\frac{g(x)^\top p_{t,c}}{\|g(x)\|_2},
 \qquad q_t(x)=\frac{A_t(x)-\mu_t^a}{\sigma_t^a},
 \label{eq:par}
\end{equation}
where $\mu_t^a$ and $\sigma_t^a$ are the mean and standard deviation of
training affinities computed with the corresponding training-time encoder for $x\in\mathcal D_t$.
Setting $K=1$ yields the best PAC accuracy in the neighborhood-size sweep
(Appendix~\ref{app:pac-topk}).
The bounded affinity correction is
\begin{equation}
 S_t^{\mathrm{PAC}}(x)=\sigma_t\tanh(q_t(x)).
 \label{eq:pac-correction}
\end{equation}
PAC adds $S_t^{\mathrm{PAC}}(x)$ to the task logit.
Affinity above the task's training mean increases the score, while affinity below
the mean decreases it.

\paragraph{FDC Framework.}
\label{sec:fusion}
After learning all $T$ tasks, FDC first applies TSF to obtain the base task logit
$\widetilde s_t(x)$ from the final adapted features $E_T(x)$.
RLC and PAC then calibrate this score using the input's compatibility with each task.
In the adapted view, they compare $E_T(x)$ with statistics collected from
$E_t(\mathcal D_t)$ when task $t$ was learned.
The pretrained view uses $E_0$ for both statistics collection and inference.
For a single view, the calibrated task logit is
\begin{equation}
 S_t(x)=\widetilde s_t(x)
       -\alpha S_t^{\mathrm{RLC}}(x)+\beta S_t^{\mathrm{PAC}}(x),
 \label{eq:score-fusion}
\end{equation}
where $\alpha$ and $\beta$ control the strengths of residual and affinity calibration.

Dual-view calibration combines corrections from the adapted and pretrained encoders
while retaining the same TSF base score:
\begin{equation}
 \begin{aligned}
 S_t^{\mathrm{dual}}(x)=\widetilde s_t(x)
 &+w_T\bigl[\beta S_t^{\mathrm{PAC},T}(x)-\alpha S_t^{\mathrm{RLC},T}(x)\bigr]\\
 &+w_0\bigl[\beta S_t^{\mathrm{PAC},0}(x)-\alpha S_t^{\mathrm{RLC},0}(x)\bigr],
 \end{aligned}
 \label{eq:dual-main}
\end{equation}
where superscripts $T$ and $0$ denote the final adapted encoder $E_T$ and the
frozen pretrained encoder $E_0$, weighted by $w_T$ and $w_0$, respectively.
Across all datasets and baseline methods, we fix the RLC/PAC correction weights and
the two view weights to one (i.e., $\alpha=\beta=w_T=w_0=1$), the PCA retained-energy
threshold to $\eta=0.75$, and the TSF attenuation strength to $\gamma=0.5$.
These shared settings avoid dataset- or baseline-specific hyperparameter tuning.

\paragraph{Plug-and-play.}
FDC can be integrated into existing continual learning methods using the shared
hyperparameter settings above, without method-specific tuning.
We organize FDC into three independently selectable components: TSF, pretrained-view
calibration ($E_0$ FDC), and adapted-view calibration ($E_T$ FDC).
Both $E_0$ FDC and $E_T$ FDC combine RLC and PAC and can be used individually or together.

%% file: sections/experiments.tex
% !TEX root = ../main.tex
\section{Evaluation}
\label{sec:experiments}
\subsection{Implementation}
\label{sec:experimental-settings}
Following the dataset setup of PILOT~\citep{sun2025pilot}, we evaluate on
CIFAR-100~\citep{krizhevsky2009cifar}, ImageNet-A~\citep{hendrycks2021imageneta},
ImageNet-R~\citep{hendrycks2021imagenetr}, CUB-200~\citep{wah2011cub}, and
OmniBenchmark~\citep{zhang2022omnibenchmark}. CIFAR-100 and CUB-200 are divided into
20 tasks with 5 and 10 classes per task, respectively. ImageNet-A, ImageNet-R, and
OmniBenchmark use 10 tasks with 20, 20, and 30 classes per task. We report final CIL
accuracy as mean $\pm$ sample standard deviation over three training seeds with a
fixed class order; gains are measured in percentage points (pp). Data partitions and
seed settings are detailed in Appendix~\ref{app:protocol}.

\paragraph{Cumulative-LoRA baseline.}
Our reference baseline, \emph{Cumulative-LoRA} (Section~\ref{sec:motivation}), trains a
new LoRA increment and head per task while freezing previous updates and heads.
The increments accumulate in a shared encoder used by all heads at inference.

All methods use ViT-B/16~\citep{dosovitskiy2021vit} with ImageNet-21K AugReg
pretraining~\citep{steiner2022vit}. LoRA runs use rank-10 key/value increments, batch
size 64, 20 main-training epochs per task, Adam, zero weight decay, and LoRA/head
learning rates of $5\times10^{-4}/5\times10^{-3}$. We reproduce each method as closely
as possible under this protocol; implementation details are provided in
Appendix~\ref{app:training}.
Each comparison with and without FDC reuses the same trained
checkpoint and test examples. Computational and memory costs are reported in
Appendix~\ref{app:cost}. Component and view ablations are reported in
Appendix~\ref{app:ablations}; average incremental accuracy for complete training trajectories is provided in Appendix~\ref{app:aia}.

\subsection{Main Results}
\label{sec:main-results}
\input{tables/benchmark}
Table~\ref{tab:current-benchmark} reports final CIL accuracy before and after FDC, with mean $\pm$ sample standard deviation over three seeds. On our reference baseline, Cumulative-LoRA, FDC improves accuracy by 4.28, 5.97, 3.58, 10.88, and 6.35 pp on CIFAR-100, ImageNet-A, ImageNet-R, CUB-200, and OmniBenchmark, respectively, without changing the trained model. Improvements are consistent across all 15 dataset--seed pairs, with the largest gain on CUB-200.

\subsection{Plug-and-Play Calibration}
\label{sec:plugin}
We apply FDC to EWC-LoRA~\citep{zheng2026ewclora}, SD-LoRA~\citep{wu2025sdlora},
SplitLoRA~\citep{qiu2026splitlora}, InfLoRA and its classifier-alignment variant
(InfLoRA+CA)~\citep{liang2024inflora},
LoRA-Subtraction~\citep{liu2025sub}, LoDA~\citep{he2026loda}, and
RSIAT~\citep{zhao2026rsiat}, using each method's trained checkpoint without retraining.
For each method, we enable all components whose standalone gains, averaged over
seeds, are positive on every dataset. The resulting configuration is shared across all
five datasets and three seeds. Appendix~\ref{app:plugin-configurations} reports the selected component configurations and their full-component counterparts, together with paired gains in Table~\ref{tab:plugin}. With selective configurations, FDC improves all 40 method--dataset
pairs by 4.39 pp on average. Enabling all components for every method yields a comparable
4.45 pp gain, with improvements in 35 of the 40 pairs and a positive five-dataset average
for every method. Thus the overall benefit persists without method-specific component
selection. FDC also improves InfLoRA+CA by 2.92 pp, supporting its compatibility with
classifier alignment under the same numerical hyperparameters.

%% file: tables/benchmark.tex
% Generated by scripts/export_merged_benchmark.py from per-seed metrics.
\begin{table}[!t]
\centering\small
\setlength{\tabcolsep}{3pt}
\renewcommand{\arraystretch}{1.05}
\caption{Final CIL accuracy (\%) before and after FDC. Bold marks the best mean per dataset.}
\label{tab:current-benchmark}
\sbox{\fdctablebox}{%
\begin{tabular}{lccccc}
\toprule
Method & CIFAR-100 & ImageNet-A & ImageNet-R & CUB-200 & OmniB. \\
\midrule
\rowcolor{pink!20}Cumulative-LoRA & \pct{84.34}{0.15} & \pct{54.11}{0.17} & \pct{75.91}{0.34} & \pct{73.55}{1.70} & \pct{67.15}{0.21} \\
\rowcolor{pink!20}\quad + FDC & \pct{88.62}{0.13} & \pct{60.08}{1.02} & \pct{79.48}{0.24} & \pct{84.44}{0.26} & \pct{73.49}{0.39} \\
\addlinespace[2pt]
EWC-LoRA & \pct{82.50}{0.26} & \pct{51.39}{0.60} & \pct{74.31}{0.33} & \pct{70.77}{0.11} & \pct{66.88}{0.37} \\
\rowcolor{blue!5}\quad + FDC & \pct{87.31}{0.43} & \pct{56.33}{0.63} & \pct{77.50}{0.29} & \pct{83.31}{0.25} & \pct{74.56}{0.30} \\
\addlinespace[2pt]
SD-LoRA & \pct{84.87}{0.85} & \pct{53.35}{0.86} & \pct{75.77}{0.23} & \pct{71.67}{1.09} & \pct{68.13}{0.19} \\
\rowcolor{blue!5}\quad + FDC & \pct{89.20}{0.25} & \pct{59.05}{0.23} & \pct{78.98}{0.28} & \pct{82.99}{0.68} & \pct{74.39}{0.61} \\
\addlinespace[2pt]
SplitLoRA & \pct{88.20}{0.28} & \pct{53.70}{0.71} & \pct{78.68}{0.13} & \pct{75.56}{0.36} & \pct{70.35}{0.38} \\
\rowcolor{blue!5}\quad + FDC & $\mathbf{89.82\!\pm\!0.12}$ & \pct{56.37}{0.44} & \pct{80.13}{0.04} & \pct{81.27}{0.15} & \pct{73.08}{0.17} \\
\addlinespace[2pt]
InfLoRA & \pct{84.82}{0.58} & \pct{51.55}{2.18} & \pct{75.06}{0.28} & \pct{72.53}{1.11} & \pct{66.71}{0.62} \\
\rowcolor{blue!5}\quad + FDC & \pct{89.03}{0.31} & \pct{58.04}{1.07} & \pct{78.81}{0.22} & \pct{83.09}{0.60} & \pct{74.36}{0.10} \\
\addlinespace[2pt]
InfLoRA+CA & \pct{84.41}{0.70} & \pct{58.77}{0.96} & \pct{76.66}{0.30} & \pct{84.34}{0.67} & \pct{71.89}{0.44} \\
\rowcolor{blue!5}\quad + FDC & \pct{88.96}{0.30} & \pct{60.90}{0.57} & \pct{79.13}{0.25} & \pct{86.57}{0.51} & $\mathbf{75.10\!\pm\!0.26}$ \\
\addlinespace[2pt]
LoRA-Subtraction & \pct{34.61}{3.91} & \pct{56.92}{0.93} & \pct{76.09}{0.18} & \pct{87.21}{0.32} & \pct{42.38}{1.56} \\
\rowcolor{blue!5}\quad + FDC & \pct{39.03}{6.65} & \pct{57.14}{1.36} & \pct{76.88}{0.32} & \pct{87.63}{0.06} & \pct{48.69}{1.41} \\
\addlinespace[2pt]
LoDA & \pct{87.41}{0.62} & \pct{62.78}{0.27} & \pct{80.69}{0.17} & \pct{77.76}{0.23} & \pct{69.83}{0.15} \\
\rowcolor{blue!5}\quad + FDC & \pct{88.96}{0.48} & $\mathbf{64.76\!\pm\!0.40}$ & $\mathbf{81.41\!\pm\!0.24}$ & \pct{82.85}{0.30} & \pct{72.77}{0.28} \\
\addlinespace[2pt]
RSIAT & \pct{54.81}{0.29} & \pct{60.87}{0.76} & \pct{60.46}{0.48} & \pct{87.59}{0.40} & \pct{61.96}{0.56} \\
\rowcolor{blue!5}\quad + FDC & \pct{66.93}{0.65} & \pct{61.73}{0.34} & \pct{66.64}{0.69} & $\mathbf{88.71\!\pm\!0.51}$ & \pct{67.42}{0.49} \\
\bottomrule
\end{tabular}}
\ifdim\wd\fdctablebox>\linewidth
\resizebox{\linewidth}{!}{\usebox{\fdctablebox}}
\else\usebox{\fdctablebox}\fi
\end{table}

%% file: sections/limitations.tex
% !TEX root = ../main.tex
\section{Limitations}
\label{sec:limitations}
Although \method{} improves inference without task identity, a substantial gap to
the Given Task ID accuracy remains.
Our calibration study is restricted to methods that share a single adapted encoder
across task heads, with Cumulative-LoRA as the reference baseline. We have not studied
how to integrate \method{} into inference over multiple candidate experts or task-specific subnetworks, as used by GR-LoRA~\citep{lin2026grlora} and TPL~\citep{lin2024tpl}, although they incur substantial additional computational and storage costs.

%% file: sections/conclusion.tex
% !TEX root = ../main.tex
\section{Conclusion}
\method{} improves task routing in class-incremental learners sharing an adapted encoder
through subspace filtering, residual likelihood, and prototype affinity, without additional
training. Its consistent gains on Cumulative-LoRA and broad applicability to existing methods
show that feature-statistic calibration helps learners better use their retained knowledge
when task identity is unavailable.

%% file: sections/ai.tex
% !TEX root = ../main.tex
\section*{AI Use Statement}
AI-assisted tools were used to help organize the manuscript, draft and edit text and equations, inspect related-work sources, and prepare tables from existing experiment outputs. Scientific interpretations, bibliographic accuracy, and the reported results remain the authors' responsibility.

%% file: sections/appendix.tex
% !TEX root = ../main.tex
\section{Experimental Setup and Evaluation Protocol}
\label{app:implementation}
\subsection{Datasets, Hyperparameters, and Metrics}
\label{app:protocol}
\label{app:diagnostics}
Table~\ref{tab:training} summarizes the class-disjoint, equal-sized task partitions and processed image counts.
CIFAR-100 uses its canonical split; the other datasets follow the processed PILOT
splits~\citep{sun2025pilot}, including a CUB-200 split that differs from the official one.

\begin{table}[!htb]
\centering\small
\caption{Dataset partitions and task sequences.}
\label{tab:training}
\begin{tabular}{lrrrrr}
\toprule
Dataset & Classes & Tasks & Classes/task & Train & Test\\
\midrule
CIFAR-100 & 100 & 20 & 5 & 50,000 & 10,000\\
ImageNet-A & 200 & 10 & 20 & 5,981 & 1,519\\
ImageNet-R & 200 & 10 & 20 & 24,000 & 6,000\\
CUB-200 & 200 & 20 & 10 & 9,430 & 2,358\\
OmniBenchmark & 300 & 10 & 30 & 89,697 & 5,985\\
\bottomrule
\end{tabular}
\end{table}

Task identity is available during training and statistics collection, but not at test time.

Table~\ref{tab:app-settings} lists the common training and FDC hyperparameters. Method-specific mechanisms are retained as described below; binary component switches are separate from the numerical hyperparameters.

\begin{table}[!htb]
\centering\small
\caption{Common training and FDC hyperparameters.}
\label{tab:app-settings}
\begin{tabular}{ll}
\toprule
Setting & Value\\
\midrule
Main-training epochs per task / batch size & 20 / 64\\
Optimizer / weight decay & Adam / 0\\
LoRA learning rate / head learning rate & $5\times10^{-4}$ / $5\times10^{-3}$\\
LoRA rank / adapted projections & 10 / key and value\\
Training seeds / class-order seed & $1,2,3$ / 1993\\
\midrule
PCA retained energy $\eta$ & 0.75\\
TSF attenuation $\gamma$ & 0.5\\
RLC and PAC weights $(\alpha,\beta)$ & $(1,1)$\\
Enabled-view weights $(w_T,w_0)$ & $(1,1)$ for dual-view calibration\\
PAC prototype neighbors $K$ & 1\\
\bottomrule
\end{tabular}
\end{table}

We report mean final CIL accuracy and sample standard deviation over three seeds, with
Raw/FDC gains paired within each seed and datasets weighted equally. Given Task ID
(Oracle) restricts prediction to the true task's classes using the same final model;
task-routing accuracy measures correct task assignments.

\subsection{Backbone and Training Settings}
\label{app:training}
All methods use ImageNet-21K AugReg-pretrained ViT-B/16 with 768-dimensional CLS features.
Cumulative-LoRA trains only the current task's rank-10 key/value increments and head;
all accumulated increments are active at inference.

\paragraph{Method-specific implementation.}
EWC-LoRA~\citep{zheng2026ewclora} estimates per-sample Fisher importance with
regularization strength $10^7$ and accumulation factor one.
SplitLoRA~\citep{qiu2026splitlora} retains gradient splitting with key/value updates.
InfLoRA~\citep{liang2024inflora} retains SVD initialization, historical-space projection,
and its native cosine schedule; the learning-rate trajectories are therefore not identical
across methods. InfLoRA+CA adds five classifier-alignment epochs on later tasks, using
Adam with learning rate $5\times10^{-3}$ and zero weight decay.
LoRA-Subtraction~\citep{liu2025sub} uses its projected Adam update.
LoDA~\citep{he2026loda} retains its two branches with optional classifier alignment disabled.
SD-LoRA~\citep{wu2025sdlora} uses its standard forward inference and native historical state.
RSIAT~\citep{zhao2026rsiat} retains its native adapter, paired drift estimation, and ten
classifier-calibration epochs on later tasks, with synthetic-feature batch size 256 and
three warmup epochs.
FDC preserves each method's classifier, including cosine scoring for LoDA and RSIAT,
EWC-LoRA's head-input transform, and LoRA-Subtraction's nearest-class-mean scoring.
Raw/FDC comparisons share checkpoints and test examples; calibration statistics use
training images with evaluation preprocessing and no gradients.

\subsection{Feature-Level Routing Diagnostic}
\label{app:feature-diagnostic}
For Figure~\ref{fig:routing-osp}, we measure the contribution of feature components
outside a head's own-training PCA subspace to its predicted-class logit:
\begin{equation}
 A(x)=\frac{w^\top(I-\Pi)h}{\sigma},
 \label{eq:motivation-residual-contribution}
\end{equation}
where $h=E_T(x)$, $\Pi$ projects onto the head's own-training principal subspace,
$w$ is the weight vector of its predicted class before calibration, and $\sigma$
is its own-training maximum-logit standard deviation.
We average this contribution separately over own-task and foreign-task test inputs
for each head. Positive values indicate that the out-of-subspace component raises
the predicted-class logit; negative values indicate that it lowers the logit.

\section{FDC Algorithm}
\label{app:method-details}
\label{app:algorithm}
Algorithm~\ref{alg:fdc} summarizes FDC for the Cumulative-LoRA reference. The two phases
collect statistics after each task and calibrate predictions at test time. The baseline
training procedure is unchanged.

\begin{algorithm}[!htb]
\caption{Full dual-view FDC: statistics collection and inference}
\label{alg:fdc}
\small
\begin{algorithmic}[1]
\REQUIRE Task stream $\{\mathcal D_t\}_{t=1}^T$, pretrained encoder $E_0$, baseline learner;
$\eta=0.75$, $\gamma=0.5$, unit correction and view weights.
\STATE \textbf{Training-time statistics:} initialize $\mathcal F^A,\mathcal F^0\gets\emptyset$.
\FOR{$t=1,\ldots,T$}
  \STATE Train task $t$ with the baseline learner to obtain $E_t,W_t,b_t$.
  \STATE Extract $G_t^A=E_t(\mathcal D_t)$ and $G_t^0=E_0(\mathcal D_t)$ without gradients.
  \FOR{$v\in\{A,0\}$}
    \STATE Fit the task mean, PCA basis, class prototypes, and own residual/affinity moments.
    \STATE For each $j<t$, append standardized residuals $u_j^v(G_t^v)$ to $\mathcal F^v$
    with task-pair labels $(j,t)$.
  \ENDFOR
  \STATE Fold TSF into $\widetilde W_t=(1-\gamma)W_t+\gamma(W_tU_t^A)(U_t^A)^\top$.
  \STATE Compute $\sigma_t=\Std_{g\in G_t^A}[\max_c(\widetilde W_tg+b_t)_c]$
  from the cached features.
  \STATE Retain task statistics; release current-task images and extracted features.
\ENDFOR
\STATE If $T>1$, for each view fit $(\mu_F^v,V_F^v)$ by Eq.~\eqref{eq:foreign} and
$\tau^v$ as the median absolute foreign log-likelihood ratio; discard $\mathcal F^v$.
\STATE \textbf{Test-time inference:} compute $g^A=E_T(x)$ and $g^0=E_0(x)$ once,
sharing frozen backbone weights; cache each view's unit-normalized feature.
\FOR{$t=1,\ldots,T$}
  \STATE Compute $\widetilde z_t=\widetilde W_tg^A+b_t$ and
  $(\widetilde s_t,c_t)=(\max_c\widetilde z_{t,c},\arg\max_c\widetilde z_{t,c})$.
  \STATE Compute PAC and RLC in both views using stored statistics and $\sigma_t$;
  set RLC to zero when $T=1$ (no foreign-reference fit).
  \STATE Combine the corrections with $\widetilde s_t$ by Eq.~\eqref{eq:dual-main} to obtain $S_t$.
\ENDFOR
\STATE Select $\hat t=\arg\max_t S_t$ and return the global class indexed by $(\hat t,c_{\hat t})$.
\end{algorithmic}
\end{algorithm}

\subsection{Training-Time Statistics Collection}
After learning task $t$, we extract $E_t(\mathcal D_t)$ and $E_0(\mathcal D_t)$ without
gradients and fit each view's task mean, PCA basis, class prototypes, and own-task moments.
The adapted-view statistics are indexed by $A$ in Algorithm~\ref{alg:fdc}. In the
checkpoint-based evaluation, they are reconstructed from each task's corresponding
checkpoint and training partition, using evaluation preprocessing.

For the foreign residual reference, current-task features are evaluated against stored
statistics of earlier tasks $j<t$. Let $\mu_{jt}$ and $V_{jt}$ be the population mean
and variance of the resulting standardized residuals for task pair $(j,t)$. With
$\mathcal P=\{(j,t):j<t\}$, each view fits
\begin{equation}
 \mu_F=\frac1{|\mathcal P|}\sum_{(j,t)\in\mathcal P}\mu_{jt},\qquad
 V_F=\frac1{|\mathcal P|}\sum_{(j,t)\in\mathcal P}
       \bigl[V_{jt}+(\mu_{jt}-\mu_F)^2\bigr].
 \label{eq:foreign}
\end{equation}
The foreign moments weight task pairs equally. The scale $\tau$ is the median absolute
log-likelihood ratio over their concatenated residual samples. Current-task features can
be released after statistics collection, and foreign residual scalars can be discarded
after fitting $\tau$.

\subsection{Test-Time Calibration}
Each test input is encoded once by $E_T$ and, when enabled, once by $E_0$. TSF transforms
the linear classifier weights as in Eq.~\eqref{eq:osp}; PAC and RLC use the original
encoder features and the stored task statistics. Both views use the same adapted-head
own-training score scale $\sigma_t$. The calibrated task logit determines the head, and
that head's local argmax determines the class.

When TSF is disabled, both the local logits and their own-training score scales revert
to Raw. For normalized or nonlinear baseline heads, the evaluator applies the TSF linear
map at the head input and retains the method's subsequent scoring operations.
Affinity, residual, and score moments use population standard deviations. Their scales,
foreign variances, and LLR scales are floored at $10^{-6}$; feature normalization and
centered-energy denominators use $10^{-12}$ safeguards. With one task, RLC is skipped.

Store $U_t^v$ rather than dense projectors $\Pi_t^v$; apply projections as
$(hU_t^v)(U_t^v)^\top$. Adapted-only calibration omits all $E_0$ computations and
statistics. Calibration using only $E_0$ still requires $E_T$ for class prediction.

\section{Computational and Memory Costs of FDC}
\label{sec:efficiency}
\label{app:cost}
We measure Cumulative-LoRA on all five datasets using the seed-1 configurations, one
NVIDIA A100 80GB PCIe GPU, FP32, and batch size 64. Statistics fitting uses four CPU
threads. Data loading uses two workers, except for OmniBenchmark, which uses eight.

\subsection{Training-Time Costs}
Table~\ref{tab:app-cost} compares baseline optimization with optimization plus FDC
statistics collection, reporting the baseline cost and the increase due to FDC.
We replay 20 training epochs per task from the corresponding
preceding checkpoint. Since FDC leaves optimization unchanged, the combined time is
the sum of the measured optimization and statistics-collection times; combined peak
GPU memory is the maximum across these sequential phases. The frozen encoder is kept
off GPU during optimization. Timings include data loading and statistics fitting but
exclude checkpoint loading, test evaluation, and artifact serialization. Peak GPU memory counts allocated tensors. FDC storage
counts factorized PCA bases, prototypes, feature means, and scalar statistics, excluding
model weights.
\input{tables/cost_collection}

\subsection{Inference-Time Costs}
Table~\ref{tab:app-latency} compares Cumulative-LoRA with single-view ($E_T$) and
dual-view ($E_T,E_0$) FDC on the same final checkpoint. We time three repeats of
20 batches after 10 warm-up batches, using 64 test images preloaded on GPU and
synchronizing at each timing boundary. Image loading is excluded. Latency columns
report total batch time; memory columns report the baseline peak and increases over it.
TSF is folded into the heads before timing; Raw retains no FDC-specific GPU tensors.
Dual reuses the frozen backbone weights for its two sequential forward passes,
disabling LoRA for $E_0$. Each view's normalized feature is reused across task heads.
Parentheses in the latency columns give increases over the baseline. Single adds 3.7--5.9 MiB and 0.7--1.7\% latency. Dual adds 7.8--12.1 MiB, with latency 1.88--1.90 times the baseline.
\input{tables/cost_inference}

TSF and adapted-view RLC share a stored PCA basis. For feature dimension $d$, total
class count $C$, and per-task/view rank $r_t^v$, statistical storage scales as
$\mathcal O(d\sum_{v,t}r_t^v+Cd+Td)$, excluding encoder and classifier weights.

\section{FDC Component Ablation Studies}
\label{app:ablations}
All ablations use the same Cumulative-LoRA checkpoints and test examples. Unless varied,
we use $\eta=0.75$, $\gamma=0.5$, $K=1$, and unit correction and view weights.

\label{sec:components}
\input{tables/ablation}
Table~\ref{tab:current-ablation} reports Raw accuracy and paired gains over Raw for all eight component combinations, with dual-view RLC and PAC. Values are means and sample standard deviations over three seeds; gain deviations are computed after subtracting Raw within each seed. Each component alone improves Raw on all five datasets, and the full
combination performs best. Component gains interact: adding PAC to TSF+RLC changes
the gain on ImageNet-R by only 0.01 pp, compared with 3.83 pp on CUB-200.

\paragraph{Single- and dual-view calibration.}
\label{sec:pretrained-routing-view}
\label{app:views}
The final two rows of Table~\ref{tab:current-ablation} retain TSF and use PAC and RLC
from either $E_T$ or $E_0$ alone. TSF acts on the adapted head; $E_0$-only calibration still uses $E_T$ for class prediction. Combining both views improves over adapted-view
calibration by 1.44, 2.05, and 1.90 pp on CIFAR-100, CUB-200, and OmniBenchmark,
respectively; ImageNet-A and ImageNet-R change little.

\section{Hyperparameter Analysis of TSF and PAC}
\label{app:hyperparameters}
We analyze the TSF attenuation and PCA retained-energy threshold, followed by the PAC neighborhood size, using five datasets and three training seeds.
\subsection{TSF Hyperparameters}
\label{app:tsf-pca-sensitivity}
We vary $\gamma\in\{0,0.25,0.375,0.5,0.625,0.75,1\}$ at $\eta=0.75$ and $\eta\in\{0.5,0.65,0.70,0.75,0.80,0.85,0.95\}$ at $\gamma=0.5$, using full dual-view FDC on the same Cumulative-LoRA checkpoints for all five datasets and three seeds. Other settings remain fixed. Statistics and calibration scales are fitted for each setting using the same causal collection protocol, without retraining the models. Curves show the three-seed mean with one sample standard deviation shaded. Vertical axes are scaled independently, and dotted lines mark the shared defaults.
\begin{figure}[!htbp]
\centering
\includegraphics[width=\linewidth]{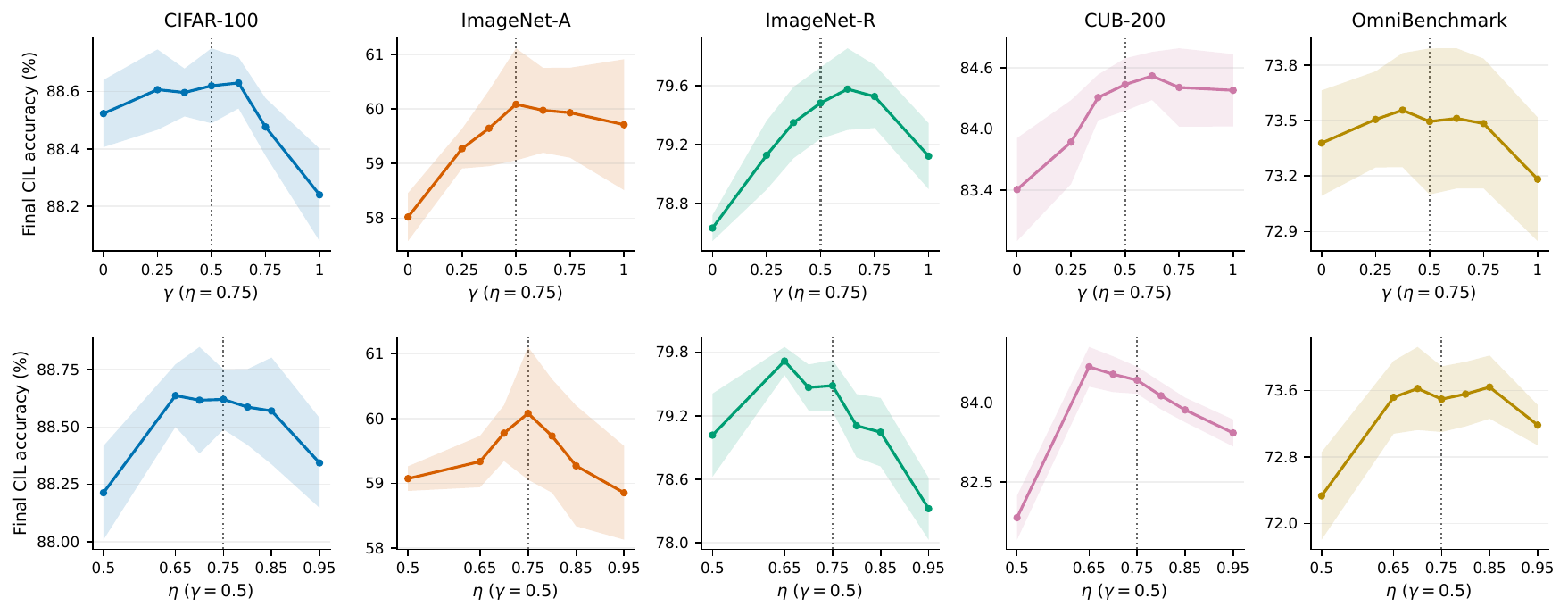}
\caption{TSF sensitivity: $\gamma$ (top) and $\eta$ (bottom), one dataset per column.}
\label{fig:tsf-hyperparameters}
\end{figure}

\paragraph{TSF attenuation.}
In Figure~\ref{fig:tsf-hyperparameters}, the default $\gamma=0.5$ averages 77.22\% across datasets, exceeding both no attenuation (76.39\%) and full attenuation (76.93\%); $\gamma=0.75$ is only 0.06 pp lower.

\paragraph{PCA retained energy.}
The default $\eta=0.75$ averages 77.22\%, close to 77.18\% at $\eta=0.65$. Dataset-specific maxima differ. Both defaults were fixed before this analysis, without dataset-specific selection; these results do not establish universal optimality.

\subsection{PAC Neighborhood Size}
\label{app:pac-topk}
\label{sec:proximity}
For each candidate task, PAC averages its $K$ highest class-prototype cosine similarities:
\begin{equation}
 A_{t,K}(x)=\frac1K\sum_{c\in\mathcal N_{t,K}(x)}
               \frac{g(x)^\top p_{t,c}}{\|g(x)\|_2}.
\end{equation}
At $K=1$, this reduces to Eq.~\eqref{eq:par}. We compare two reference distributions for standardizing $A_{t,K}$: \emph{Distributional matching} uses own-task training statistics recomputed for each $K$, whereas \emph{Nearest} uses statistics across candidate tasks for the current input. Both apply the same $\sigma_t\tanh(q_{t,K})$ correction; prototypes, TSF, dual-view RLC, head-score scales, and local predictions remain fixed.

\begin{figure}[H]
\centering
\includegraphics[width=0.80\linewidth]{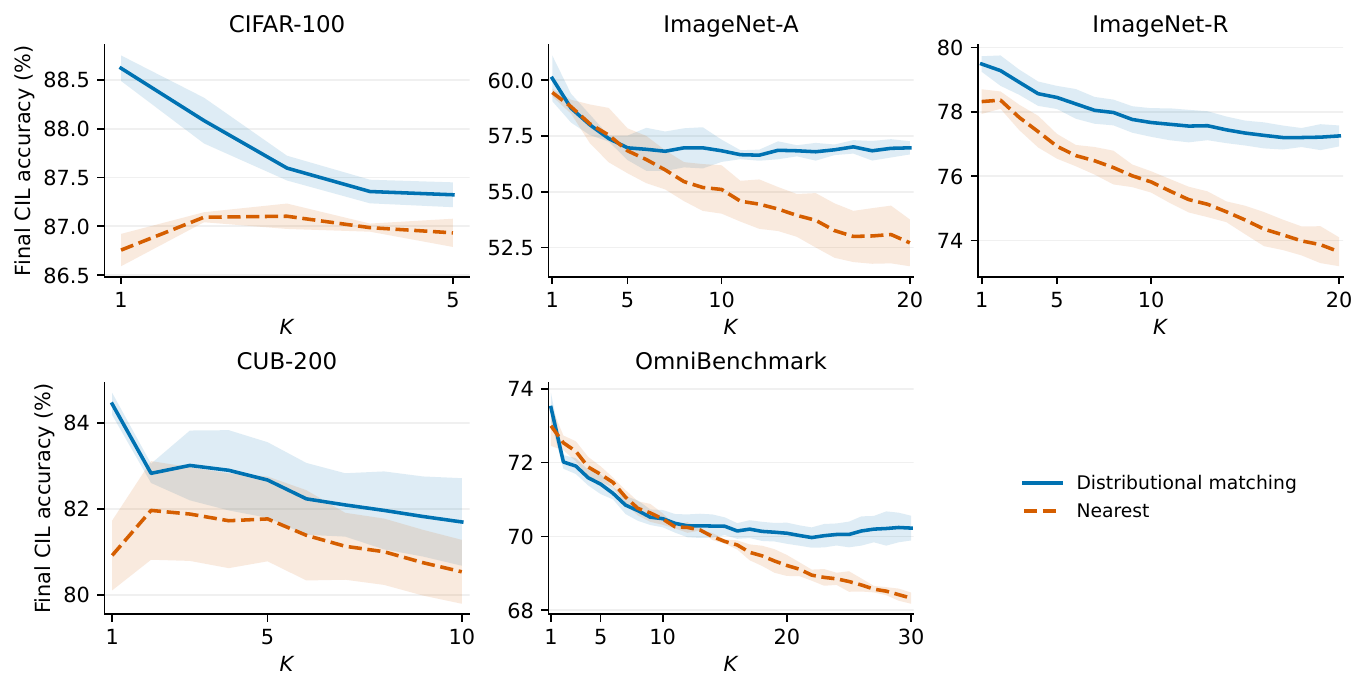}
\caption{PAC accuracy across neighborhood sizes.}
\label{fig:app-pac-k}
\end{figure}

Figure~\ref{fig:app-pac-k} sweeps all feasible $K$, showing three-seed means and sample standard deviations. Matching performs best at $K=1$ and exceeds Nearest on all five datasets there. Nearest can lead at larger $K$, especially on OmniBenchmark, but remains below the default matching result there.

\section{Average Incremental Accuracy}
\label{app:additional-results}
\label{app:aia}
For CIL accuracy $A_k$ over all encountered classes after task $k$, average incremental accuracy is $\mathrm{AIA}=T^{-1}\sum_{k=1}^T A_k$. Table~\ref{tab:current-aia} reports the mean and sample standard deviation over three seeds for the compared shared-encoder methods. FDC is evaluated at the final checkpoint, so its AIA is not reported.
\input{tables/aia}

\section{Plug-and-Play Component Configurations}
\label{app:plugin-configurations}
Table~\ref{tab:plugin} gives the configurations used in Table~\ref{tab:current-benchmark}. Following Section~\ref{sec:plugin}, each method enables components with positive standalone mean gains on all five datasets, then uses this configuration across datasets and seeds. This selection uses the reported evaluation results; Cumulative-LoRA enables all components.

The TSF, $E_0$, and $E_T$ switches denote filtering, pretrained-view calibration, and adapted-view calibration (1: enabled; 0: disabled). Each calibration view includes RLC and PAC. Where selection disables a component, a second row reports full FDC. Gains are paired with Raw within each seed and reported as mean $\pm$ sample standard deviation; Avg. weights datasets equally.
\input{tables/plugin}

%% file: tables/cost_collection.tex
\begin{table}[!htb]
\centering\small
\setlength{\tabcolsep}{4pt}
\caption{Training-stage time, GPU memory, and FDC storage.}
\label{tab:app-cost}
\begin{tabular*}{\linewidth}{@{\extracolsep{\fill}}lrrrrr@{}}
\toprule
& \multicolumn{2}{c}{Time (min)} & \multicolumn{2}{c}{Peak GPU memory (GiB)} & FDC storage\\
\cmidrule(lr){2-3}\cmidrule(lr){4-5}
Dataset & Baseline & + FDC & Baseline & + FDC & (MiB)\\
\midrule
CIFAR-100 & 132.2 & +5.1 & 6.51 & +0.00 & 11.86 \\
ImageNet-A & 18.7 & +1.1 & 6.51 & +0.00 & 10.05 \\
ImageNet-R & 64.8 & +5.1 & 6.51 & +0.00 & 11.30 \\
CUB-200 & 29.3 & +3.0 & 6.51 & +0.00 & 7.68 \\
OmniB. & 239.3 & +9.2 & 6.51 & +0.00 & 12.03 \\
\bottomrule
\end{tabular*}
\end{table}

%% file: tables/cost_inference.tex
% Source: build/inference-cost-shared-20260922/*/result.json.
\begin{table}[!htb]
\centering\small
\setlength{\tabcolsep}{3pt}
\caption{Inference latency and GPU memory at batch size 64.}
\label{tab:app-latency}
\begin{tabular*}{\linewidth}{@{\extracolsep{\fill}}lrrrrrr@{}}
\toprule
& \multicolumn{3}{c}{Total latency (ms/batch)} & \multicolumn{3}{c}{Peak GPU memory (MiB)}\\
\cmidrule(lr){2-4}\cmidrule(lr){5-7}
Dataset & Baseline & Single & Dual & Baseline & $\Delta$ Single & $\Delta$ Dual\\
\midrule
CIFAR-100 & 170.5 & 173.3 (+1.7\%) & 323.1 (+89.5\%) & 980.9 & +4.7 & +11.9 \\
ImageNet-A & 171.0 & 172.5 (+0.8\%) & 321.5 (+88.0\%) & 981.4 & +5.0 & +10.1 \\
ImageNet-R & 170.5 & 171.7 (+0.7\%) & 320.1 (+87.7\%) & 981.4 & +5.9 & +11.3 \\
CUB-200 & 171.3 & 174.1 (+1.6\%) & 324.2 (+89.3\%) & 981.4 & +3.7 & +7.8 \\
OmniB. & 169.9 & 171.3 (+0.8\%) & 319.2 (+87.9\%) & 982.0 & +5.7 & +12.1 \\
\bottomrule
\end{tabular*}
\end{table}

%% file: tables/ablation.tex
% Values rounded once from current_bs64_osp75_analysis_20260915/summary.json.
% Export: scripts/export_ablation_deltas.py; paired differences before rounding.
\begin{table}[t]
\centering\small
\setlength{\tabcolsep}{4pt}
\caption{Component and view ablations: Raw accuracy (\%) and gains over Raw (pp).}
\label{tab:current-ablation}
\sbox{\fdctablebox}{%
\begin{tabular}{lccccc}
\toprule
Method & CIFAR-100 & ImageNet-A & ImageNet-R & CUB-200 & OmniB. \\
\midrule
Raw & \pct{84.34}{0.15} & \pct{54.11}{0.17} & \pct{75.91}{0.34} & \pct{73.55}{1.70} & \pct{67.15}{0.21} \\
TSF & \pct{+0.85}{0.22} & \pct{+3.29}{0.43} & \pct{+1.45}{0.28} & \pct{+2.19}{0.84} & \pct{+0.37}{0.32} \\
Dual PAC & \pct{+2.89}{0.21} & \pct{+2.96}{0.23} & \pct{+1.63}{0.21} & \pct{+8.14}{1.64} & \pct{+4.83}{0.26} \\
Dual RLC & \pct{+2.76}{0.14} & \pct{+3.45}{1.25} & \pct{+2.34}{0.20} & \pct{+4.95}{0.83} & \pct{+5.02}{0.12} \\
TSF + Dual PAC & \pct{+3.21}{0.27} & \pct{+4.56}{0.42} & \pct{+2.63}{0.25} & \pct{+9.44}{1.46} & \pct{+4.90}{0.23} \\
TSF + Dual RLC & \pct{+3.21}{0.16} & \pct{+5.66}{0.86} & \pct{+3.57}{0.14} & \pct{+7.05}{0.84} & \pct{+5.03}{0.19} \\
Dual PAC + Dual RLC & \pct{+4.18}{0.21} & \pct{+3.91}{0.40} & \pct{+2.73}{0.26} & \pct{+9.85}{1.36} & \pct{+6.23}{0.27} \\
Full (Dual FDC) & \pct{+4.28}{0.13} & \pct{+5.97}{0.91} & \pct{+3.58}{0.13} & \pct{+10.88}{1.48} & \pct{+6.35}{0.31} \\
\midrule
FDC ($E_T$ calibration) & \pct{+2.84}{0.35} & \pct{+5.92}{0.59} & \pct{+3.58}{0.32} & \pct{+8.84}{1.57} & \pct{+4.44}{0.12} \\
FDC ($E_0$ calibration) & \pct{+3.76}{0.12} & \pct{+4.50}{0.56} & \pct{+2.09}{0.12} & \pct{+8.71}{0.92} & \pct{+5.70}{0.04} \\
\bottomrule
\end{tabular}}
\ifdim\wd\fdctablebox>\linewidth
  \resizebox{\linewidth}{!}{\usebox{\fdctablebox}}
\else\usebox{\fdctablebox}\fi
\end{table}

%% file: tables/aia.tex
% Source: artifacts/manuscript/posthoc-routing-20260918/results.md
% SHA256: 7763f22192227efe4bb8f9a7504565caa734de7cfa10cbdc7e14612f6a1542c9
% REV-20260918 T03: Refresh complete AIA records and retain the same architecture scope as the main table.
\begin{table}[!htb]
\centering\small
\setlength{\tabcolsep}{3pt}
\caption{Average incremental accuracy (\%), mean $\pm$ standard deviation.}
\label{tab:current-aia}
\sbox{\fdctablebox}{%
\begin{tabular}{lccccc}
\toprule
Method & CIFAR-100 & ImageNet-A & ImageNet-R & CUB-200 & OmniB. \\
\midrule
Cumulative-LoRA & \pct{90.56}{0.16} & \pct{63.20}{0.35} & \pct{82.09}{0.31} & \pct{82.76}{1.36} & \pct{77.17}{0.13} \\
EWC-LoRA & \pct{88.88}{0.40} & \pct{61.39}{0.40} & \pct{81.14}{0.24} & \pct{80.95}{0.54} & \pct{76.58}{0.27} \\
SD-LoRA & \pct{90.81}{0.44} & \pct{62.81}{0.64} & \pct{81.94}{0.08} & \pct{80.96}{1.79} & \pct{77.53}{0.17} \\
SplitLoRA & \pct{92.49}{0.29} & \pct{63.83}{0.63} & \pct{84.48}{0.18} & \pct{82.55}{0.26} & \pct{78.66}{0.30} \\
InfLoRA & \pct{90.80}{0.23} & \pct{61.67}{1.84} & \pct{81.65}{0.16} & \pct{82.65}{1.17} & \pct{76.23}{0.38} \\
InfLoRA+CA & \pct{91.33}{0.12} & \pct{68.42}{0.57} & \pct{82.36}{0.14} & \pct{90.43}{0.37} & \pct{80.36}{0.09} \\
LoRA-Subtraction & \pct{63.06}{2.27} & \pct{67.03}{0.86} & \pct{81.94}{0.25} & \pct{92.08}{0.22} & \pct{67.07}{0.63} \\
LoDA & \pct{92.51}{0.35} & \pct{72.80}{0.61} & \pct{86.63}{0.05} & \pct{85.89}{0.18} & \pct{78.98}{0.06} \\
RSIAT & \pct{77.13}{0.53} & \pct{72.73}{0.36} & \pct{77.43}{0.09} & \pct{92.86}{0.17} & \pct{77.34}{0.24} \\
\bottomrule
\end{tabular}}
\ifdim\wd\fdctablebox>\linewidth
\resizebox{\linewidth}{!}{\usebox{\fdctablebox}}
\else\usebox{\fdctablebox}\fi
\end{table}

%% file: tables/plugin.tex
% Source: artifacts/manuscript/posthoc-routing-20260918/results.md
% SHA256: 7763f22192227efe4bb8f9a7504565caa734de7cfa10cbdc7e14612f6a1542c9
% Full rows: validated osp_dual_fdc results in build/all-on-20260922/summary.json.
% Avg. uses unrounded gains; SD-LoRA is 6.16486 pp.
\begin{table}[!htbp]
\centering\small
\setlength{\tabcolsep}{2pt}
\renewcommand{\arraystretch}{1.10}
\caption{FDC component configurations and paired accuracy gains (pp).}
\label{tab:plugin}
\sbox{\fdctablebox}{%
\begin{tabular}{lccccccccc}
\toprule
Baseline & TSF & $E_0$ & $E_T$ & CIFAR-100 & ImageNet-A & ImageNet-R & CUB-200 & OmniB. & Avg. \\
\midrule
EWC-LoRA & 1 & 1 & 1 & +\pct{4.81}{0.41} & +\pct{4.94}{0.63} & +\pct{3.19}{0.23} & +\pct{12.54}{0.35} & +\pct{7.68}{0.17} & +6.63 \\
\noalign{\vskip 2pt}\hdashline[1pt/2pt]\noalign{\vskip 2pt}
SD-LoRA & 1 & 1 & 1 & +\pct{4.33}{0.60} & +\pct{5.71}{0.95} & +\pct{3.21}{0.15} & +\pct{11.32}{0.88} & +\pct{6.26}{0.72} & +6.16 \\
\noalign{\vskip 2pt}\hdashline[1pt/2pt]\noalign{\vskip 2pt}
\multirow{2}{*}{SplitLoRA} & 0 & 0 & 1 & +\pct{1.62}{0.16} & +\pct{2.68}{0.55} & +\pct{1.46}{0.09} & +\pct{5.71}{0.32} & +\pct{2.74}{0.21} & +2.84 \\
 & 1 & 1 & 1 & $+1.76\!\pm\!0.22$ & $+1.95\!\pm\!0.44$ & $+0.75\!\pm\!0.14$ & $+9.10\!\pm\!0.35$ & $+3.48\!\pm\!0.23$ & +3.41 \\
\noalign{\vskip 2pt}\hdashline[1pt/2pt]\noalign{\vskip 2pt}
InfLoRA & 1 & 1 & 1 & +\pct{4.21}{0.28} & +\pct{6.50}{1.15} & +\pct{3.74}{0.09} & +\pct{10.56}{0.51} & +\pct{7.66}{0.72} & +6.53 \\
\noalign{\vskip 2pt}\hdashline[1pt/2pt]\noalign{\vskip 2pt}
InfLoRA+CA & 1 & 1 & 1 & +\pct{4.55}{0.94} & +\pct{2.13}{0.47} & +\pct{2.47}{0.09} & +\pct{2.23}{0.47} & +\pct{3.22}{0.50} & +2.92 \\
\noalign{\vskip 2pt}\hdashline[1pt/2pt]\noalign{\vskip 2pt}
\multirow{2}{*}{\shortstack[l]{LoRA-\\Subtraction}} & 0 & 0 & 1 & +\pct{4.41}{2.98} & +\pct{0.22}{0.55} & +\pct{0.79}{0.16} & +\pct{0.42}{0.36} & +\pct{6.31}{0.15} & +2.43 \\
 & 1 & 1 & 1 & $+6.28\!\pm\!2.77$ & $-2.09\!\pm\!0.21$ & $-0.32\!\pm\!0.20$ & $-0.99\!\pm\!0.28$ & $+11.09\!\pm\!0.23$ & +2.80 \\
\noalign{\vskip 2pt}\hdashline[1pt/2pt]\noalign{\vskip 2pt}
\multirow{2}{*}{LoDA} & 0 & 0 & 1 & +\pct{1.55}{0.17} & +\pct{1.98}{0.17} & +\pct{0.72}{0.21} & +\pct{5.09}{0.07} & +\pct{2.94}{0.28} & +2.45 \\
 & 1 & 1 & 1 & $+1.07\!\pm\!0.17$ & $+1.71\!\pm\!0.37$ & $+0.11\!\pm\!0.05$ & $+5.08\!\pm\!0.17$ & $+3.53\!\pm\!0.37$ & +2.30 \\
\noalign{\vskip 2pt}\hdashline[1pt/2pt]\noalign{\vskip 2pt}
\multirow{2}{*}{RSIAT} & 0 & 1 & 0 & +\pct{12.12}{0.45} & +\pct{0.86}{0.46} & +\pct{6.18}{0.23} & +\pct{1.12}{0.19} & +\pct{5.46}{0.24} & +5.15 \\
 & 1 & 1 & 1 & $+13.07\!\pm\!0.19$ & $-0.09\!\pm\!0.83$ & $+6.68\!\pm\!0.55$ & $-0.18\!\pm\!0.10$ & $+4.71\!\pm\!0.53$ & +4.84 \\
\bottomrule
\end{tabular}}
\ifdim\wd\fdctablebox>\linewidth
\resizebox{\linewidth}{!}{\usebox{\fdctablebox}}
\else\usebox{\fdctablebox}\fi
\end{table}